\documentclass{article}
\pdfoutput=1

     \usepackage[final]{neurips_2022}

\usepackage[utf8]{inputenc} % allow utf-8 input
\usepackage[T1]{fontenc}    % use 8-bit T1 fonts
\usepackage{hyperref}
\usepackage{url}            % simple URL typesetting
\usepackage{booktabs}       % professional-quality tables
\usepackage{amsfonts}       % blackboard math symbols
\usepackage{nicefrac}       % compact symbols for 1/2, etc.
\usepackage{microtype}      % microtypography
\usepackage{xcolor}         % colors
\usepackage{subfigure}
\usepackage{multirow}
\usepackage{graphicx}
\usepackage{amsmath}
\usepackage{amsthm}
\usepackage{amsfonts,amssymb}
\usepackage{bm}
\usepackage{enumitem}
\usepackage{makecell}
\theoremstyle{plain}

\theoremstyle{definition}

\title{Transposed Variational Auto-encoder with Intrinsic Feature Learning for Traffic Forecasting}

\author{%
Leyan Deng$^{\dag\S *}$, Chenwang Wu$^{\dag }$\thanks{These authors contributed equally to this work.} , Defu Lian$^{\dag}$, Min Zhou$^{\S}$ \\
$^{\dag}$ School of Data Science, University of Science and Technology of China\\
$^{\S}$ Huawei Noah's Ark Lab\\
\texttt{\{dleyan, wcw1996\}@mail.ustc.edu.cn}, \texttt{liandefu@ustc.edu.cn}, \texttt{zhoumin27@huawei.com}\\
}

\begin{document}
\maketitle
\begin{abstract}
  In this technical report, we present our solutions to the Traffic4cast 2022 core challenge and extended challenge. In this competition, the participants are required to predict the traffic states for the future 15-minute based on the vehicle counter data in the previous hour. Compared to other competitions in the same series, this year focuses on the prediction of different data sources and sparse vertex-to-edge generalization. To address these issues, we introduce the Transposed Variational Auto-encoder (TVAE) model to reconstruct the missing data and Graph Attention Networks (GAT) to strengthen the correlations between learned representations. We further apply feature selection to learn traffic patterns from diverse but easily available data. 
  Our solutions have ranked first in both challenges on the final leaderboard. The source code is available at \url{https://github.com/Daftstone/Traffic4cast}.
\end{abstract}

\section{Introduction}
The growing populations and vehicles of cities bring a challenge to efficient and sustainable mobility \cite{zheng2014urban,zheng2015detecting}. Therefore, Intelligent Transportation \cite{zhang2011data,bibri2017smart}, especially traffic state predictions, are of great social and environmental value \cite{pmlr-v123-kreil20a,zhang2017deep}. The Institute of Advanced Research in Artificial Intelligence (IARAI) has hosted three competitions at NeurIPS 2019, 2020, and 2021 \cite{pmlr-v123-kreil20a,pmlr-v133-kopp21a,pmlr-v176-eichenberger22a}. In the previous competitions, the organizer provided large-scale datasets and designed significant tasks to advance the application of AI to forecasting traffic. 
Traffic4cast 2022 provides large-scale road dynamic graphs of different cities and asks participants to make predictions on different data sources. This competition requires models that have the ability to generalize vertex data to the traffic states of the entire graph. Here sparse loop counter data and prediction for the entire city will drive efficient low-barrier traffic forecasting to possible. In this work, we present similar frameworks for two challenges. Our contributions are as follows.
\begin{itemize}
\item In order to alleviate the extreme sparsity of vehicle counter data, we build a Transposed Variational Auto-encoder (TVAE) model \cite{kingma2013auto} to reconstruct the missing data. Compared with VAE, which unreasonably reconstructs all missing values to the same value, it can reconstruct more meaningful values.
\item Since there is few information in one-hour dynamic data, we further consider static graph structure and time information to capture traffic patterns, which are useful but easily available.
\item Inspired by the high intrinsic dependencies within road graphs, we apply Graph Attention Networks (GAT) to enhance the connections between learned representations. 
\end{itemize}

\section{Methodology}

We solve both tasks using a similar strategy. Specifically, they both include (1) data preprocessing, (2) vehicle counter reconstruction, (3) feature representation, and (4) feature fusion and learning. Below we describe each module in detail. Table \ref{tab: notation} shows the key notations used in this paper.

\begin{table}[htbp]
	\centering
	\caption{Summary of key notations}
	\small
	\setlength{\tabcolsep}{0.007\linewidth}{
	\begin{tabular}{cc}
		\hline\noalign{\smallskip}
		Notation & Definition \\
		\noalign{\smallskip}\hline\noalign{\smallskip}
        $V$     & the set of nodes \\
        $E$     & the set of edges \\
        $S$     & the set of super-segments \\
		$X$     & Spatially sparse vehicle counters $X\in (\mathbb{R}\cup NaN)^{|V|\times 4}$ \\
        $M$     & It indicates whether the value in $X$ is missing (=0) or not (=1)\\ 
		$\mathcal{U}_s$     & Node embedding $\mathcal{U}_s\in\mathbb{R}^{|V|\times d}$ \\
        $\mathcal{U}_d$     & Reconstructed vehicle counters $\mathcal{U}_d\in\mathbb{R}^{|V|\times 4}$\\
        $\mathcal{V}_e$     & Explicit edge features extracted from dataset\\
        $\mathcal{V}_i$     & Implicit edge embedding $V_i\in \mathbb{R}^{|E|\times d}$\\
        $\mathcal{U}_w$     & Week embedding $\mathcal{U}_w\in \mathbb{R}^{7\times d}$\\
        $\mathcal{U}_t$     & Time embedding $\mathcal{U}_t\in \mathbb{R}^{96\times d}$\\
        $A_{SV}$    & Adjacency matrix $A_{SV}\in\{0,1\}^{|S|\times|V|}$ of super-segments and nodes\\
        $A_{SE}$    & Adjacency matrix $A_{SE}\in\{0,1\}^{|S|\times|E|}$ of super-segments and edges\\
		$N_{ind}$ & Node index $N_{ind}\in \mathbb{N}_+^{E\times 2}$, where $N_{ind}^i$ denotes that two nodes of $i$-th edge are $N_{ind}^{i,0}$ and $N_{ind}^{i,1}$ \\
		\noalign{\smallskip}\hline
	\end{tabular}%
}
	\label{tab: notation}%
\end{table}%

\subsection{Data Preprocessing}
\label{sec: preprocess}
Traffic4cast 2022 provides datasets of three cities, including London, Madrid, and Melbourne \footnote{\url{https://github.com/iarai/NeurIPS2022-traffic4cast}}. In each city, given a directed road graph $G(V, E, S)$, where $V$, $E$, and $S$ are sets of nodes, edges, and super-segments, respectively. There are two tracks in this competition, and the organizers provide the same inputs for both challenges, which are vehicle counters in 15-minute aggregated time bins for one prior hour to the prediction time slot. The participants in the core challenge are asked to classify each edge into red, yellow, or green. In the extended challenge, they are asked to predict the travel time of each super-segment. Then the input shape in both challenges is ${|V|\times 4}$, the output shape in the core challenge is $|E|\times 3$, and the output in the extended task is $|S| \times 1$.  

We first compute the mean value and standard deviation using the entire training dataset of each city, then we adopt z-score normalization on all inputs. To amplify the variability between volume values, we hard clip the maximum value of three processed datasets as the maximum value (i.e., 23.91) in London. This is because the maximum values of these two datasets are much larger than 23.91, but the number of these extreme values is very small. If clipping is not performed, most of the normalized values will be indistinguishable, increasing the difficulty of feature learning. Finally, we fill the missing data with the minimum value of each dataset. 

\subsection{Vehicle Counter Reconstruction}
\begin{figure*}[ht]
	\centering
	\includegraphics[width=0.7\linewidth]{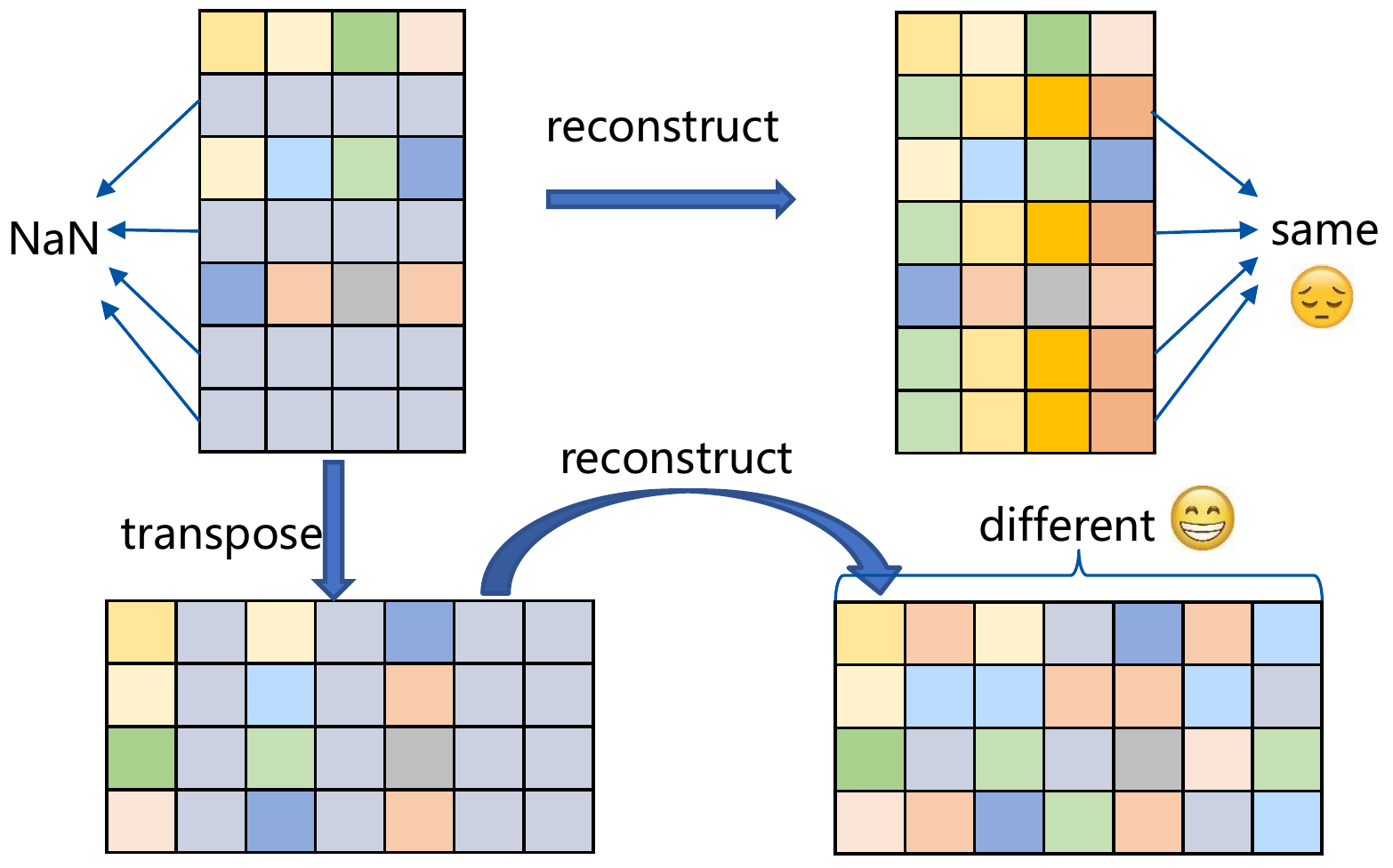}
	\caption{Motivation for using a transposed VAE. Top: the disadvantage of traditional VAE, which causes NaN to be reconstructed to a uniform value. Bottom: the transposed VAE. Since the reconstruction of NaN values is related to other non-NaNs of this sample (Note that there are only 4 samples), this makes the reconstructed values almost never the same.}
	\label{fig: reconstruct}
\end{figure*}

Even if we preprocess the data well, these tasks are still challenging. This is because it is impossible for us to obtain the traffic information of each intersection, which is reflected in the fact that there are a large number of nodes without data in the dataset. Therefore, the natural idea is to reconstruct the missing data of these nodes.

Here, we choose the Variational Auto-encoder (VAE) as our reconstruction model, but there still have problems. In Section \ref{sec: preprocess}, we filled NaNs with the minimum value. Then, as long as the nodes are missing, the preprocessed features of these nodes will be the same, and in turn, the outputs to these nodes will be the same for any reconstruction model. Formally, for any $i, j$, if $X_i=X_j=NaN$, then for any VAE $f_v(\cdot)$, we have $f_v(X_i)=f_v(X_j)$. An example is shown in the top half of Fig. \ref{fig: reconstruct}. Since the node features denote the intersection traffic in the previous hour, obviously, they will not be the same.

To this end, we propose Transposed Variational Auto-encoder (TVAE). Before reconstruction, we normalize the inputs into $[0, 1]$ using min-max normalization and finally restore the outputs of the reconstruction model to the original range. 
Specifically, we transpose the counter matrix to $X^T$, and the shape of the counter matrix $X^T$ becomes $4\times N$. As shown in the lower half of Fig. \ref{fig: reconstruct}. It can be seen that at this time, these four samples (length $N$) are basically not the same at this time, so TVAE can play its role in reconstructing the vehicle values of all nodes from the sparse counter data. We define the reconstructed matrix of $X$ as $f_v(X^T)^T$. Note that we only reconstruct nodes with missing values and keep all nodes with values, so the reconstructed features $\mathcal{U}_d=M\odot X+(1-M)\odot f_v(X^T)^T$, where $\odot$ denotes the Hadamard product.

\subsection{Feature Representation}
Only the vehicle counter is not enough, node or edge features are also important for prediction tasks. In this section, we introduce several classes of features, including static and dynamic features of nodes, static and dynamic features of edges, and temporal features, to enhance the learned representation. It is worth noting that our prediction task focuses on edges (single edge or super segment), so although node features are learned, they are ultimately used for edge prediction tasks.

\subsubsection{Congestion Prediction Task}
\label{sec: cc}
\textbf{Node features.} The features of nodes are crucial for crowding prediction as well as super-segment speed prediction. In the previous section, we reconstructed the intersection's vehicle counter for each time period. But this only represents the dynamic features of the intersection, and the static features of the intersection are also important. For example, the number of paths at the intersection, the time setting of traffic lights, and the street's prosperity will all affect the traffic flow. This is the missing feature of the dataset. To this end, we set an embedding $\mathcal{U}_s\in\mathbb{R}^{|V|\times d}$ to represent the intrinsic (static) features of each node (intersection).

With the dynamic features $\mathcal{U}_d$ and intrinsic features $\mathcal{U}_s$ of nodes, we use Graph Attention Network (GAT) \cite{brody2021attentive} to further strengthen the connection between nodes. Specifically, using two GATv2 layers, respectively, we obtain enhanced node representations:
$$\mathcal{U}_{d+}=f_{gat}(\mathcal{U}_d,,f_g(e)), \mathcal{U}_{s+}=f_{gat}(\mathcal{U}_s,,f_g(e)),$$
where $f_g(e)$ represents the use of a fully-connected layer to extract edge features $e$ as the weight of the GATv2 layer. Based on the enhanced node representation, for each edge i, we can respectively get four node embeddings $\mathcal{U}_{d+}^{N_{ind}^{i,0}}$, $\mathcal{U}_{d+}^{N_{ind}^{i,1}},\mathcal{U}_{s+}^{N_{ind}^{i,0}}$, $\mathcal{U}_{s+}^{N_{ind}^{i,1}}$ corresponding to this edge, and the we use the concatenate operator and two fully-connected layers $f_{N1}(\cdot), f_{N2}(\cdot)$ to get the node dynamic and intrinsic features of edge $i$, that is, $f_{N1}([\mathcal{U}_{d+}^{N_{ind}^{i,0}}, \mathcal{U}_{d+}^{N_{ind}^{i,1}}])$, $f_{N2}([\mathcal{U}_{s+}^{N_{ind}^{i,0}}, \mathcal{U}_{s+}^{N_{ind}^{i,1}}])$.

\textbf{Edge features.} For the edge attribute provided from the dataset, we select seven features, including $speed\_kph$, $parsed\_maxspeed$, $length\_meters$, $counter\_distance$, $importance$, $highway$, and $oneway$. We apply the one-hot encoding on discrete features and min-max normalization on continuous features to get the explicit features $\mathcal{V}_e$. Then we use a fully-connected layer $f_E(\cdot)$ to get its high-level representation $f_E(\mathcal{V}_e)$.

Similar to the processing of nodes, we also use an embedding $\mathcal{V}_i\in\mathbb{R}^{|E|\times d}$ to represent the implicit features of each edge. This can be used to represent unknown road quality, inherent foot traffic, etc.

\begin{figure*}[ht]
	\centering
	\includegraphics[width=1.\linewidth]{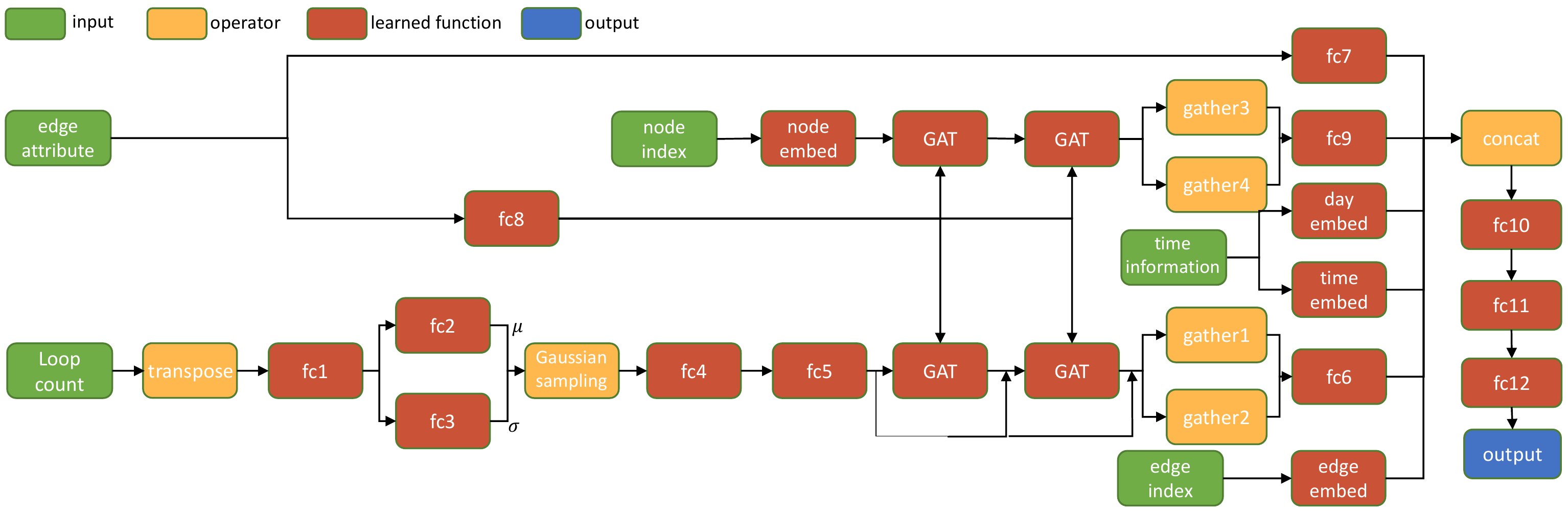}
	\caption{Model architecture in the congestion prediction task.}
	\label{fig: cc-framework}
\end{figure*}

\textbf{Temporal feature.} Temporal information is also another important factor in measuring traffic congestion. For example, the traffic pressure on weekends is obviously not as high as that on weekdays, and the traffic during commuting hours is obviously higher than that in other hours. To do this, we extract the current time from counters\_daily\_by\_node.parquet, which can be obtained by matching the current counters of all nodes with all counters. We construct two embeddings, $\mathcal{U}_w\in\mathbb{R}^{7\times d}$ and $\mathcal{U}_t\in\mathbb{R}^{96\times d}$, which represent the intrinsic features of the week and the time, respectively. Then, for the week $i$ and time $j$, we get $\mathcal{U}_w^i$ and $\mathcal{U}_t^j$ respectively, and then copy them $|E|$ times to get the current temporal feature of all edges, denoted as $\mathcal{V}_w\in\mathbb{R}^{|E|\times d}$ and $\mathcal{V}_t\in\mathbb{R}^{|E|\times d}$.

\subsubsection{Super-Segment Speeds Prediction Task}
\label{sec: eta}
The features used in the super-segment speed prediction task are similar to those used in the congestion prediction task. One difference is that the congestion prediction task is to predict each edge, and here is to predict each super-segment. Therefore, we aggregate the node features and edge features in the super-segment. Assuming that $A_{SV}$ is the adjacency matrix of the super-segment and the node, and $A_{SE}$ is the adjacency matrix of the super-segment and the edge, then we can get the dynamic and intrinsic node features of the super-segment as:
$$\mathcal{U}_{Sd}=A_{SV}\cdot \mathcal{U}_{d+},\ \ \mathcal{U}_{Ss}=A_{SV}\cdot \mathcal{U}_{s+}.$$
The explicit and implicit edge features of the super-segment as follows:
$$\mathcal{V}_{Se}=A_{SE}\cdot \mathcal{V}_e,\ \ \mathcal{V}_{Si}=A_{SE}\cdot \mathcal{V}_i.$$

In addition, like the intrinsic features of nodes and the implicit features of edges, we also use an embedding $S$ to represent the intrinsic features of super-segments.

\subsection{Feature Fusion and learning}
\textbf{Congestion Prediction Task.} We fuse the three groups of features in Section \ref{sec: cc}. Here we just use a simple concatenate operator to get the final feature $x_c$, that is,
$$x_c=[f_{N1}([\mathcal{U}_{d+}^{N_{ind}^{i,0}}, \mathcal{U}_{d+}^{N_{ind}^{i,1}}]), f_{N2}([\mathcal{U}_{s+}^{N_{ind}^{i,0}}, \mathcal{U}_{s+}^{N_{ind}^{i,1}}]),f_E(V_e),\mathcal{V}_i,\mathcal{V}_w,\mathcal{V}_t]$$
Furthermore, we use three fully connected layers to get the final congestion prediction $f_c(x_c)\in\mathbb{R}^{|E|\times 3}$.

For model training, there are two losses here: (1) Reconstruction loss. We select those non-empty nodes and minimize the Mean Squared Error before and after reconstruction, that is $\mathcal{L}_r=\frac{1}{|N'|}\sum_{i\in N'}(\mathcal{U}_d^i-x^i)^2$; (2) Classification loss. Due to the imbalance of congestion classes, we choose weighted cross-entropy loss to alleviate the training difficulty caused by data imbalance, that is, $\mathcal{L}_{wc}=\frac{1}{|E'|}\sum_{i\in E'}w_{y_i}y\log (f_c(x_c))$. Finally, the training loss is the sum of these two parts:
$$\mathcal{L}_c=\mathcal{L}_r+\mathcal{L}_{wc}.$$

\begin{figure*}[ht]
	\centering
	\includegraphics[width=1.\linewidth]{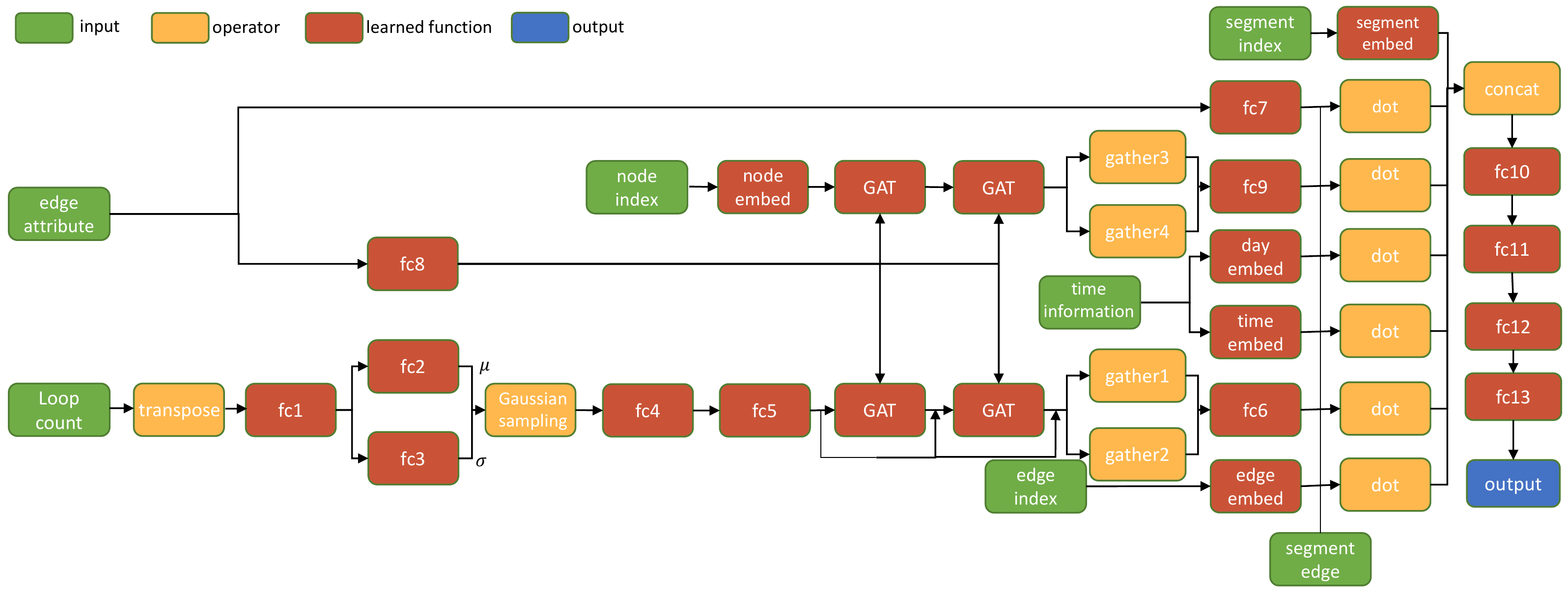}
	\caption{Model architecture in the super-segment speeds prediction task.}
	\label{fig: eta-framework}
\end{figure*}

\textbf{Super-Segment Speeds Prediction Task.} We fuse the four groups of features, including node features, edge features, temporal features, and super-segment features. We also use a simple concatenate operator to get the final feature $x_s$, that is,
$$x_s=[\mathcal{U}_{Sd},\mathcal{U}_{Ss},V_{Se},V_{Si},V_i,V_w,V_t,S].$$
Then, we use three fully connected layers to get the final speed prediction $f_s(x_s)\in\mathbb{R}^{|S|\times 1}$.

For model training, there still have a reconstruction loss $\mathcal{L}_r=\frac{1}{|N'|}\sum_{i\in N'}(\mathcal{U}_d^i-x^i)^2$. In addition, we adopt a $L_1$ loss to fit the real speed, that is, $\mathcal{L}_{l1}=\frac{1}{|S|}\sum_{i\in S}|f_s(x_s)-s_i|$. Finally, the training loss is the sum of these two parts:
$$\mathcal{L}_s=\mathcal{L}_r+\mathcal{L}_{l1}.$$

The model architecture used by the two tasks is shown in Fig. \ref{fig: cc-framework} and Fig. \ref{fig: eta-framework}.
\section{Results}
\subsection{Experimental Settings}
For the congestion prediction task, all models are trained using the AdamW optimizer \cite{loshchilov2017decoupled} with epochs of 20, batch size of 2, a learning rate of $1e-3$, and weight decay of $1e-3$. For the super-segment speed prediction task, all models are trained using the AdamW optimizer with epochs of 50, batch size of 2, and learning rate of $1e-4$. All models are trained on a single Tesla V100.

We present several key components used in our models:
\begin{itemize}
\item \textbf{Global normalization}. We normalize the input of the reconstructed model to $[0,1]$ using the global maxima and minima (i.e., the extreme value of the data at all times) and restore the output of the reconstructed model based on the global maxima and minima. Without global normalization, we compute the maxima and minima value using the current input.
\item \textbf{Dropout}. We use Dropout \cite{srivastava2014dropout} on the last three fully-connected layers to alleviate overfitting, where drop ratio $p=0.2$.
\item \textbf{Noise}. Inspired by denoising Auto-encoders, we scale the input of the reconstruction model to perform data augmentation, and the scaling factor range is $[0.8,1.2]$.
\item \textbf{Week} and \textbf{Time}. As introduced in Section \ref{sec: cc}, we use temporal features $\mathcal{U}_w$ and $\mathcal{U}_t$ to assist the prediction task.
\item \textbf{5-folds}. We divide the dataset into 5 mutually exclusive subsets and then train each of the five models to average the prediction results. 
\item \textbf{Average}. The results of the last k epochs are averaged. Here it is only applied in the super-segment speed prediction task, and k is set to 10.
\item \textbf{Segment conv}. In the super-segment speed prediction task, we put the concatenated feature $x_c$ into a GATv2 layer and then concatenate with the fully-connected layer.
\end{itemize}

\subsection{Congestion Prediction Task}
\label{sec: result-cc}
\begin{table}[htbp]
  \centering
  \caption{Performance in the congestion prediction task.}
  \setlength{\tabcolsep}{0.007\linewidth}{
    \begin{tabular}{c|cccccc|c}
    \hline\noalign{\smallskip}
    model & Global normalization & Dropout & Noise & Week  & Time   & 5-folds & Test Score \\
    \noalign{\smallskip}\hline\noalign{\smallskip}
    1     & \checkmark     &   &    &       & \checkmark       &       & 0.8511 \\
    2     & \checkmark     &   &    &       & \checkmark      & \checkmark     & 0.8461 \\
    3     &       &   &    & \checkmark     & \checkmark       &       & 0.8519 \\
    4     &       &   &    & \checkmark     & \checkmark       & \checkmark     & 0.8472 \\
    5     & \checkmark     & \checkmark  &   & \checkmark     & \checkmark          &       & 0.8508 \\
    6     & \checkmark     & \checkmark  &   & \checkmark     & \checkmark          & \checkmark     & 0.8455 \\
    7     & \checkmark     & \checkmark  & \checkmark  & \checkmark         & \checkmark     &       & 0.8501 \\
    8     & \checkmark     & \checkmark  & \checkmark  & \checkmark         & \checkmark     & \checkmark     & 0.8446 \\
    \noalign{\smallskip}\hline\noalign{\smallskip}
    Ensemble &       &   &    &       &             &       & 0.8431 \\
    \noalign{\smallskip}\hline
    \end{tabular}%
    }
  \label{tab:cc}%
\end{table}%
\begin{figure*}[ht]
	\centering
	\includegraphics[width=1.\linewidth]{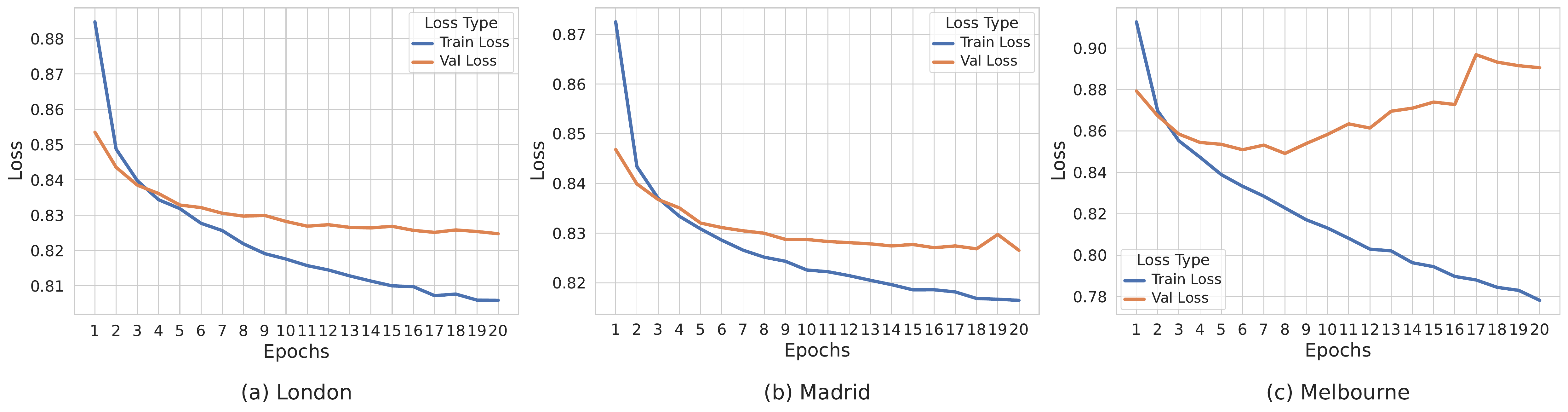}
	\caption{The loss curve of the training set and the validation set during the training process in the core task. }
	\label{fig: cc-loss}
\end{figure*}

The congestion prediction performance of various models is shown in Table \ref{tab:cc}. First, these models show competitive results, and it is worth mentioning that we also achieved first place in the final leaderboard, which verifies the effectiveness of our models. Second, individual components show their role in predicting congestion, and we achieve the best performance when we use all of them. In addition, it can be seen that the performance of the model using 5-folds is significantly improved. Finally, we found a small performance gain for all model ensembles as well. For the ensemble here, we use a weighted ensemble, that is, the model with a better score will have a larger weight.

Additionally, we plot the loss curves for the training and validation sets of the 8th model, as shown in Fig \ref{fig: cc-loss}. It can be seen that at the end of the 20th epoch, the validation set of the three cities basically reaches the optimum. However, Melbourne has an obvious over-fitting phenomenon, which we suspect is caused by the extreme imbalance of classes. This is what needs to be further considered and improved in future studies.
\subsection{Super-Segment Speeds Prediction Task}

\begin{figure*}[ht]
	\centering
	\includegraphics[width=1.\linewidth]{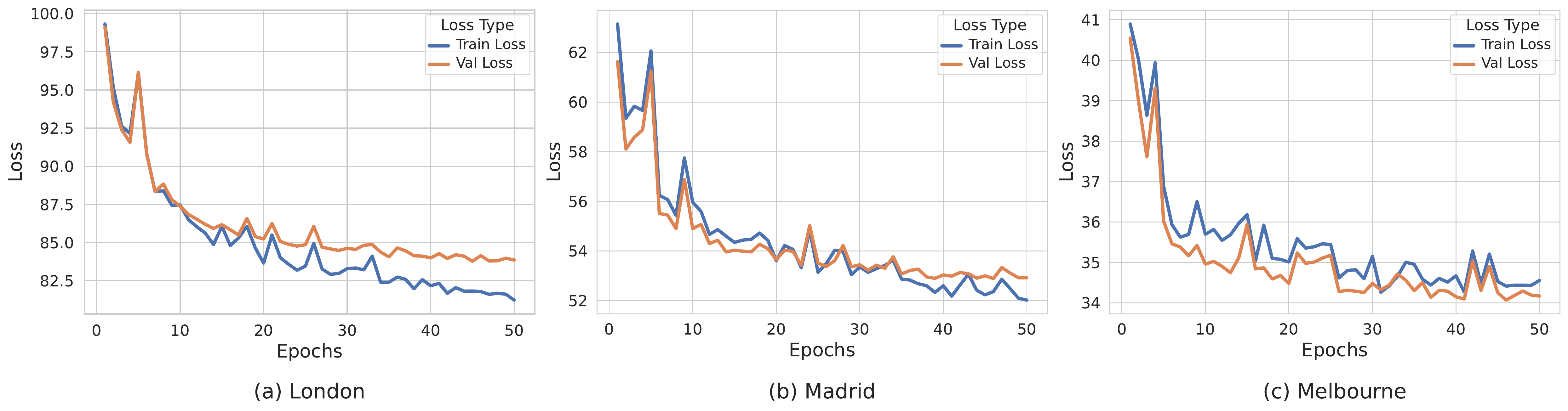}
	\caption{The loss curve of the training set and the validation set during the training process in the extended task.}
	\label{fig: eta-loss}
\end{figure*}

The super-segment speed prediction performance is shown in Table \ref{tab:eta}, and the loss curve of the best model (5th model) is illustrated in Fig. \ref{fig: eta-loss}. We can reach similar conclusions as in the congestion prediction task (Section \ref{sec: result-cc}). Besides, we also achieved first place in this track, which proves the effectiveness of our models.

\begin{table}[htbp]
  \centering
  \caption{Performance in the super-segment speeds prediction task.}
  \setlength{\tabcolsep}{0.007\linewidth}{
    \begin{tabular}{c|cccccc|c}
    \hline\noalign{\smallskip}
    Model & Global normalization & Noise & Week  & Time   & Segment conv & Average & Test score \\
    \noalign{\smallskip}\hline\noalign{\smallskip}
    1     & \checkmark     & \checkmark     & \checkmark     & \checkmark     & \checkmark & \checkmark    & 58.95 \\
    2     &       &       &       & \checkmark     & \checkmark  & \checkmark   & 59.26 \\
    3     &       &       &       & \checkmark     &   & \checkmark    & 59.34 \\
    4     & \checkmark     &       & \checkmark     & \checkmark     &    & \checkmark   & 59.01 \\
    5     &       &       &       & \checkmark     &     & \checkmark  & 59.30 \\
    \noalign{\smallskip}\hline\noalign{\smallskip}
    Ensemble &       &       &       &       &    &   & 58.50 \\
    \noalign{\smallskip}\hline
    \end{tabular}%
    }
  \label{tab:eta}%
\end{table}%

\section{Discussion}
Our method has ranked first in the core challenge and extended challenge. We first use the Transposed-VAE-based reconstruction model to obtain dense vehicle counter data. Additionally, we construct a series of features (e.g., temporal features, and graph static attributes) to alleviate the challenge of sparse data. Since it is often difficult to obtain complete loop counters in practice, the proposed feature representation enhances the practicality. Moreover, based on different types of features, we see opportunities for further work in traffic pattern analysis. In two challenges, we utilize different GNNs to capture correlations between learned representations. In particular, we use the same structure in the super-segment prediction task, and the effect is still significant, which leads to a new insight: the proposed model has good generalization and can be fine-tuned to better adapt to specific tasks.

In addition, we adopt the same network architectures and parameters for all cities. However, we observe obvious over-fitting during the training stage in the Melbourne dataset. One possible reason is the extreme imbalance of classes, which deserves more exploration in the future.

\begin{ack}
The work was supported by grants from the National Key R\&D Program of China (No. 2021ZD0111801) and the National Natural Science Foundation of China (No. 62022077).
\end{ack}

% \section*{References}
\bibliographystyle{plainnat}
\bibliography{neurips_2022}

%%%%%%%%%%%%%%%%%%%%%%%%%%%%%%%%%%%%%%%%%%%%%%%%%%%%%%%%%%%%
\end{document}